\documentclass[conference]{IEEEtran}

\IEEEoverridecommandlockouts
\usepackage{graphicx}
\usepackage[cmex10]{amsmath}
\usepackage{amsfonts}
\usepackage{amssymb}
\usepackage{amsthm}
\usepackage{booktabs}
\usepackage{algorithm}
\usepackage{algorithmic}
\usepackage[misc]{ifsym}

\usepackage{times}

\usepackage{soul}
\usepackage{dblfloatfix}    
\usepackage{url}
\usepackage{xcolor}
\usepackage{float}
\usepackage{cite}
\usepackage[pagebackref=false,breaklinks=true,colorlinks,bookmarks=false]{hyperref}

\begin{document}
\hypersetup{
  pdfinfo={
    Title={Multimodal Noisy Segmentation based fragmented burn scars identification in Amazon Rainforest},
    Author={Satyam Mohla, Sidharth Mohla, Anupam Guha, Biplab Banerjee},
    Subject={noisy, multimodal, remote, sensing, satelite, mapping, segmentation, semantic, computer, vision, burn, scar, weakly, fragmented, land, use, unet},
    Keywords={semantic, segmentation, unet, burn, scars, labelled, amazon, remote, noisy, marks, management, wildfires, sensing, identification, multimodal, training, weakly, segment, fragmented}
  }
}

\title{Multimodal Noisy Segmentation based fragmented burn scars identification in Amazon Rainforest
\thanks{$^{\star}$Equal Contribution \hspace{10pt} {\protect \includegraphics[scale=0.17]{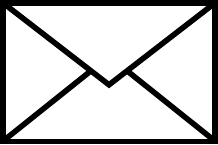}}\hspace{3pt}Corresponding Author
}
}


{\author{Satyam~Mohla$^{\star,1,3}$\href{https://orcid.org/0000-0002-5400-1127}{\protect \includegraphics[scale=0.12]{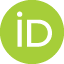}}, Sidharth~Mohla$^{\star,2,3}$\href{https://orcid.org/0000-0002-3974-1071}{\protect \includegraphics[scale=0.12]{orcid.png}}, Anupam~Guha$^{1}$, Biplab~Banerjee$^{1}$\href{https://orcid.org/0000-0001-8371-8138}{\protect \includegraphics[scale=0.12]{orcid.png}}
\\
$^{1}$Indian Institute of Technology Bombay, India\\
$^{2}$Indian Institute of Technology Hyderabad, India\\
$^{3}$Koloro Labs, India\\
{{\protect \includegraphics[scale=0.17]{drawing.pdf}}\hspace{2pt}\{satyammohla, sidmohla\}@gmail.com}
}
}


\maketitle

\begin{abstract}
Detection of burn marks due to wildfires in inaccessible rain forests is important for various disaster management and ecological studies. The fragmented nature of arable landscapes and diverse cropping patterns often thwart the precise mapping of burn scars. Recent advances in remote-sensing and availability of multimodal data offer a viable solution to this mapping problem. However, the task to segment burn marks is difficult because of its indistinguishably with similar looking land patterns, severe fragmented nature of burn marks and partially labelled noisy datasets.

In this work we present AmazonNET -- a convolutional based network that allows extracting of burn patters from multimodal remote sensing images. The network consists of UNet- a well-known encoder decoder type of architecture with skip connections commonly used in biomedical segmentation. The proposed framework utilises stacked RGB-NIR channels to segment burn scars from the pastures by training on a new weakly labelled noisy dataset from Amazonia.

Our model illustrates superior performance by correctly identifying partially labelled burn scars and rejecting incorrectly labelled samples, demonstrating our approach as one of the first to effectively utilise deep learning based segmentation models in multimodal burn scar identification.
\end{abstract}



\section{Introduction}
\IEEEPARstart{I}{n Amazonia}, fire is associated with several land-practices. Slash-and-Burn is one of the most used practices in Brazilian agriculture (as part of a seasonal cycle called \texttt{"queimada"}\cite{winklerprins2006creating}). Whether for opening and cleaning agricultural areas or renewing pastures, its importance in the agricultural chain is undeniable. Unfortunately, this is often the cause of wildfires in forests.\cite{lewis2015increasing, van2012trends,juarez2017causes}

\begin{figure}[t]
\centering
  \centerline{\includegraphics[width=9cm]{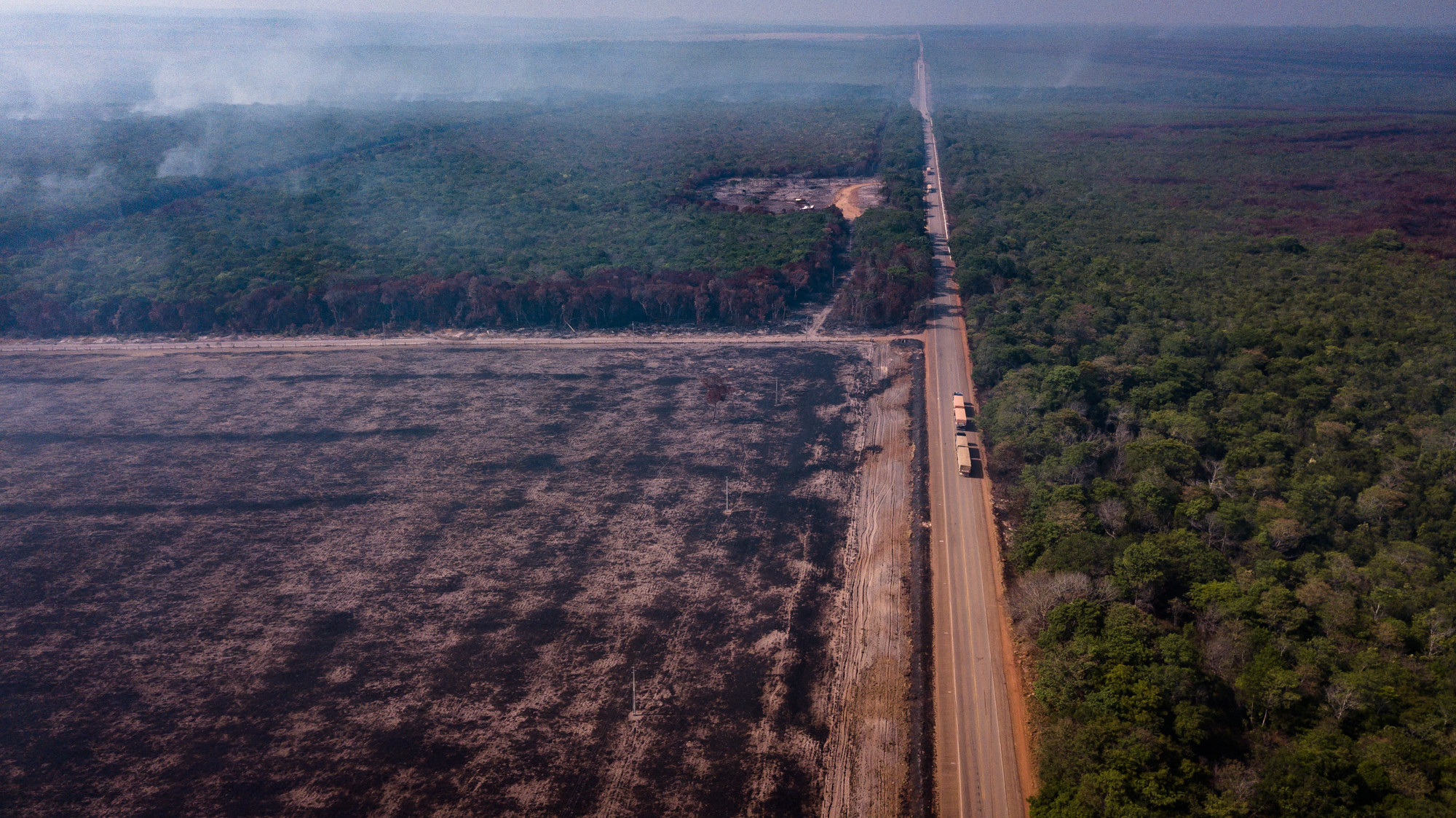}}
\caption{Overview of burnings in the vicinity of BR-163 highway, Para, northern Brazil, in Amazon region.
\textit{Taken from \cite{basso2019}}}
  \vspace{-0.5cm}
\label{fig:AmazonForest}
\end{figure}

Amazon rainforests are a major reservoir for flora and billions of tons of carbon, release of which can cause a major increase in temperatures. Recent news of wildfires in Amazon therefore, caused major uproar and concern (Fig. \ref{fig:AmazonForest}).%

Uncontrollable fires, especially in dry season, have major local \& regional impacts, leading to destruction of natural biomes, extinction of animal \& plant species, pollution, erosion and an imbalance in the carbon cycle. Such disturbances affect agricultural production as well. Thus, many environmental studies \& resources management activities require accurate identification of burned areas for monitoring the affected regions (the so-called scars from burning) spatially and temporally in order to understand and assess the vulnerability of these areas and to promote sustainable development.
\begin{figure*}[!bt]
\centering
  \centerline{\includegraphics[width=16.5cm]{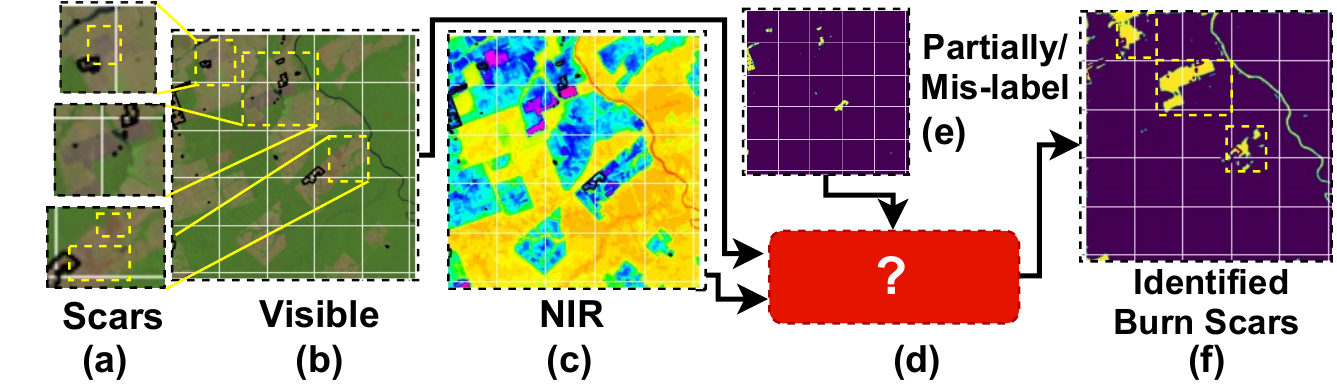}}
    \caption{Generic schematic of a multi-modal noisy, weakly labelled burn scar identification. (a) Unlabelled/Correctly labelled burn scars (b) Visible Band (c) Near Infrared Band (d) Unknown Model (e) Partially/ Noisy training labels (f) Output burn scar map}\medskip
\vspace{-10pt}
\label{fig:problem_description}
\end{figure*}
\begin{figure*}[!b]
\centering
  \centerline{\includegraphics[width=18.5cm]{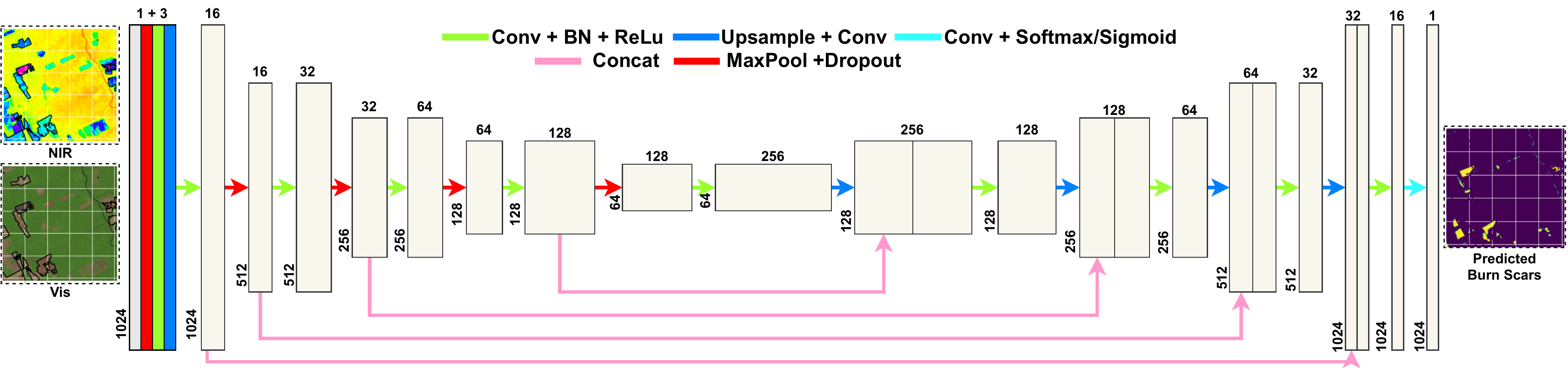}}
\caption{Architecture of U-net model: AmazonNet (presented on Amazon dataset). Input consists of 4 channel concatenated input corresponding to RGB and NIR (colormap: hsv). ReLu activation is used. 3x3 Convolution and 2x2 Max Pool are used thruout the network.}\medskip
\label{fig:AmazonNet}
\end{figure*}
Due to the large geographical extent of fires at regional and global scales and the limited accessibility of the areas affected by fire, remote sensing approaches have become cost effective alternatives in the last few years, capable of collecting burned area information at adequate spatial and temporal resolutions. Remote sensing technologies can provide useful data for fire management, estimation \& detection, fuel mapping, to post wildfire monitoring, including burn area and severity estimation\cite{lanorte2013multiscale}.

\section{Problem Statement}

Current non-deep learning methods heavily rely on domain knowledge and manual input from the user and are unable to extract the abstract representations from the data. Deep learning attempts to resolve these problems however, they remain largely neglected in burn scar prediction due to general lack of any labelled data. In this work, we leverage the recent advances in sensing leading to ubiquitous availability of multimodal data and computer vision in remote sensing to utilise noisy, weakly labelled data to identify fragmented burn scars using UNet, making our approach one of the first to utilise deep learning based segmentation models in multimodal burn scar identification. The same has been illustrated in Fig.\ref{fig:problem_description}.
\section{Related Work}
\textbf{Semantic Segmentation} \hspace{5pt} 
Semantic Segmentation is an important problem in computer vision. It involves the clustering of various pixels together if they belong to the same object class. Due to their ability to capture semantic context with precise localisation, they have been used for various applications in autonomous driving \cite{ess2009segmentation,geiger2012we,cordts2016cityscapes}, human-machine interaction \cite{oberweger2015hands}, diagnosis and detection \cite{shvets2018automatic,tschandl2019domain}, remote sensing \cite{seferbekov2018feature,zhang2019jointnet,robinson2019large} etc.
%
Before the advent of DNNs, variety of features were used for semantic segmentation, such as K-means\cite{hartigan1975clustering}, Histogram of oriented gradients\cite{dalal2005histograms}, Scale-invariant feature transform\cite{lowe2004sift,suga2008object} etc. Today, many encoder-decoder networks and their variants like SegNet \cite{badrinarayanan2017segnet}, have been proposed. Specialized applications have led to novel improvements like UNet for Medical Image Segmentation\cite{ronneberger2015u}, CRFs based networks for fine segmentation\cite{chen2018encoder}
\begin{figure*}[!hb] 
\centering
  \centerline{\includegraphics[width=18cm]{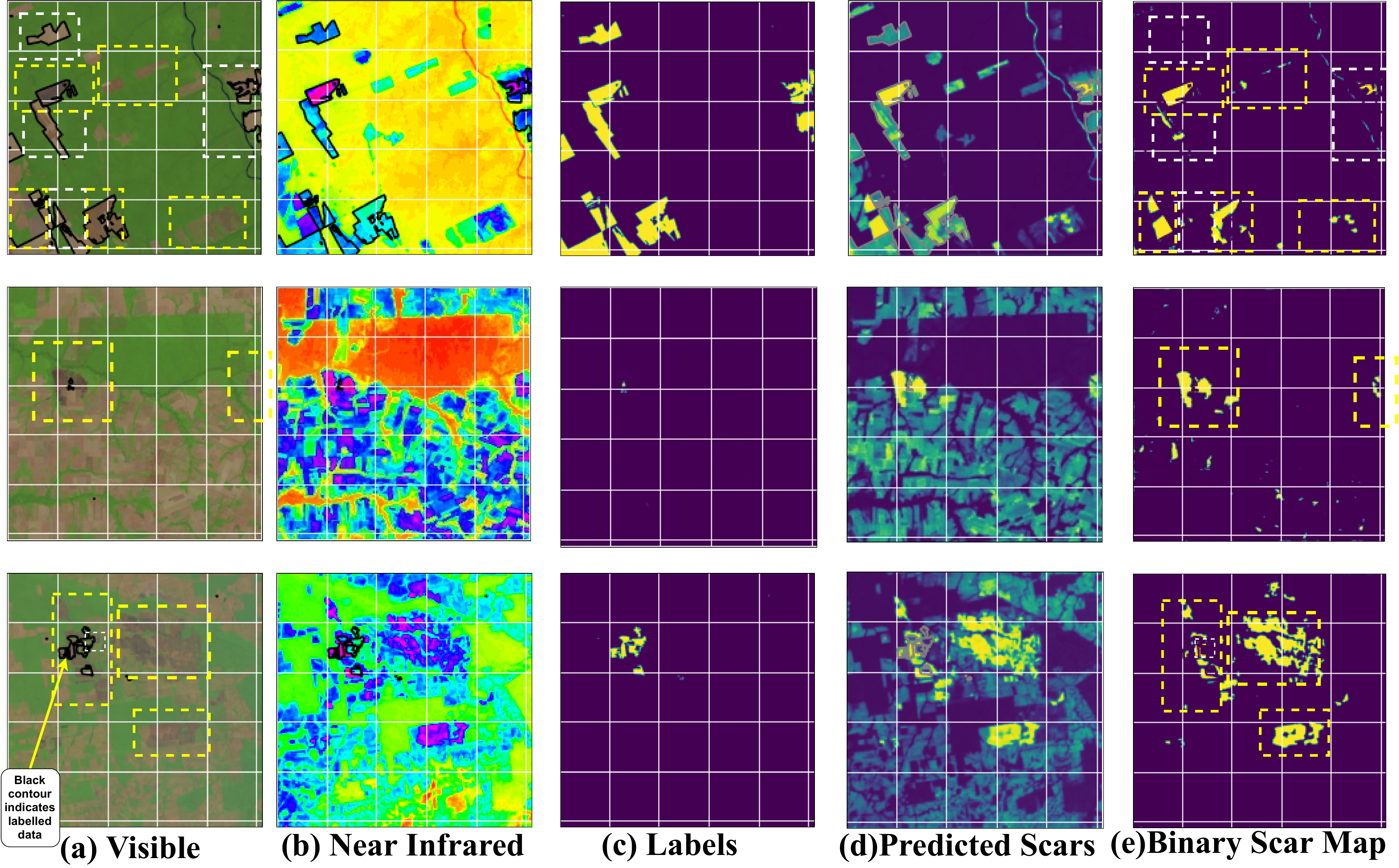}}
\caption{Results: The network correctly segments burn scars, rejecting incorrectly labelled spots and correctly identifying partially labelled or unlabelled samples. (a) Visible Channel (b) Near Infrared channel (false coloured hsv) (c) Available Labels (including partial/ incorrect labels) (d) Predicted Burn Scars (e) Binary Burn Scars. The black contour in (a) \& (b) denote contour for labelled data(c) for easy visualisation. Yellow boxes denote burn scars which are correctly labelled or unlabelled. White boxes denote mislabelled/misidentified burn scars.}\medskip
\label{fig:results}
\end{figure*}

\textbf{Multimodal data in remote sensing} \hspace{5pt}
Multimodal segmentation in remote sensing involves utilising various strategies to efficiently combine information contained in multiple modalities to generate a single, rich, fused representation beneficial for accurate land use classification. Common methods involve concatenation channels at input stage \cite{Iglovikov_2018_CVPR_Workshops}, concatenation of features extracted from unimodal networks like CNNs, as in \cite{chen2016deep,chen2017deep} to generate land mapping segmentation. Recent works involve more sophisticated ideas like `cross-attention' \cite{Mohla_2020_CVPR_Workshops} to fuse multiple modalities and generate attentive spectral and spatial representations.

\textbf{Burn Scar Identification} \hspace{5pt}
Simultaneous availability of multimodal data has led to recent advances in locating fires and in quantifying the area burned. Each modality provides discriminating information about the same geographical region, helping in mapping amidst adverse conditions like spectral confusion (like due to cloud shadowing) \& variability in burn scars making distinguishing between vegetation difficult. Majority of work done in this domain involves methods like auto-correlation \cite{lanorte2013multiscale}, self-organizing maps\cite{lasaponara2019mapping}, linear spectral mixture model\cite{barbosadetection}, SVM \cite{pereira2017burned}, random forests \cite{liu2020feasibility}. However, no recent works seem to utilise current deep learning methods like CNN or encoder-decoder models like SegNet or UNet, presumably due to lack of labelled data.

\section{Proposed Method}
{The objective of this work is to perform semantic segmentation and identify burn scars by harnessing the spatio-spectral information constituted in visible and near infrared. To accomplish this task, we consider RGB and NIR samples $\mathcal{X} = \{\textbf{x}^i_{RGB}, \textbf{x}^i_{NIR}\}_{i=1}^n$ with the ground truth $\mathcal{Y} = \{y^i\}_{i=1}^n$.

Here, $\textbf{x}^i_{RGB}$ $\in$ $\mathbb{R}^{M\times N \times B_1}$ and $\textbf{x}^i_{NIR} \in \mathbb{R}^{M\times N \times B_2}$ where, $B_1$ and $B_2$ denote the number of channels, while $n$ denotes the number of available samples. The ground-truth labels $y_i^n \in \{0,1\}$, where 0 represents `no burn scar' $\in$ \{river, green pastures, brown pastures\} and 1 represents `burn scar'. The samples are sent to the proposed AmazonNet model  which get processed as discussed ahead.
}
\subsection{Architecture}
In remote sensing, computer vision based methods are difficult to apply, due to lack of good labelled datasets, because the required data processing and labelling can only be done by field experts, making labelled data rare or unavailable. Similar problems arise in medical image segmentation, and so common approaches in remote sensing are sometimes inspired from medical segmentation domain. 
\begin{figure*}[t] 
\centering
  \centerline{\includegraphics[width=18cm]{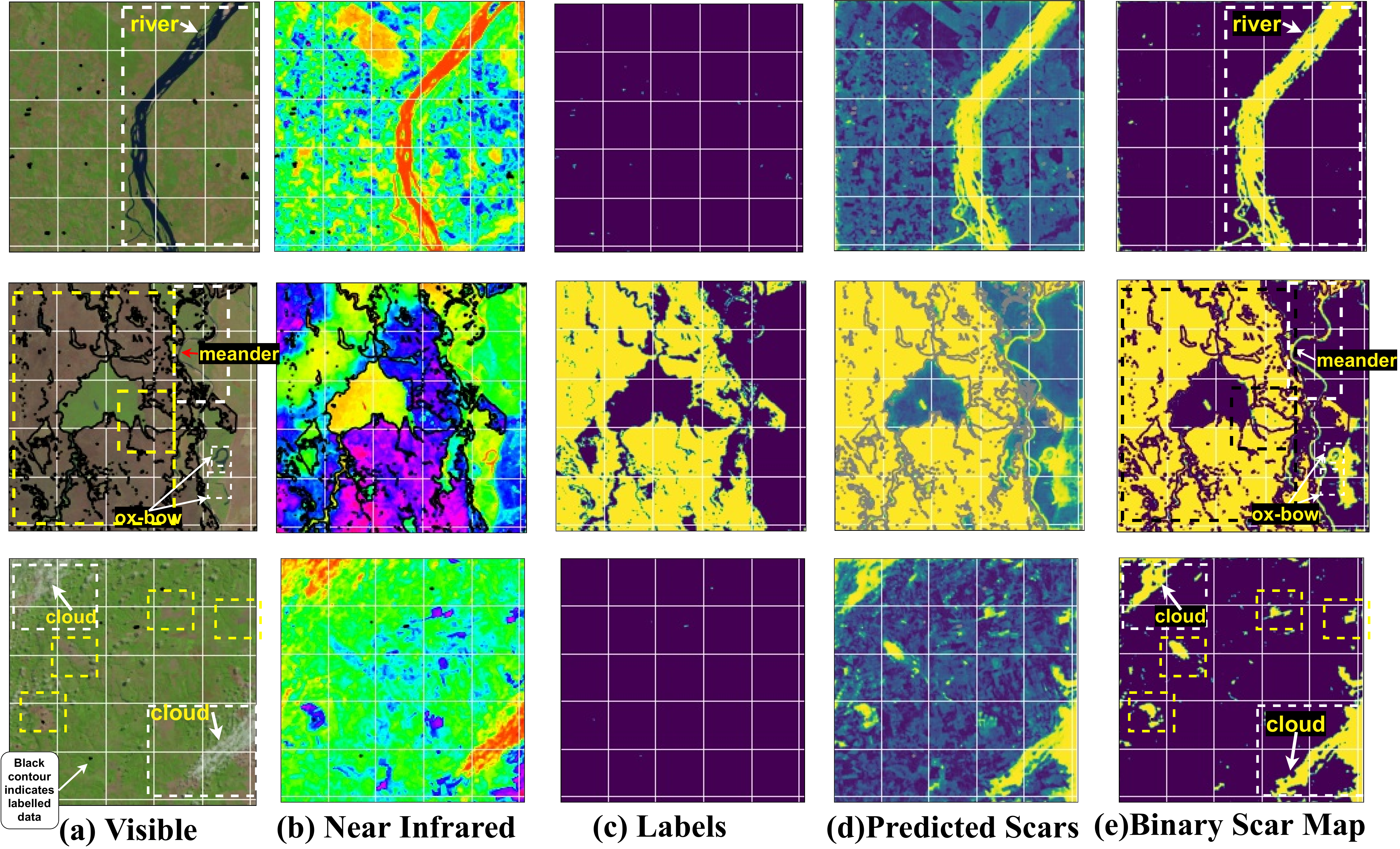}}
\caption{Minor defects in segmented burn scars: The network incorrectly segments rivers, meanders, ox-bow lakes and clouds as burn scars. The black contour in (a) \& (b) denote contour for labelled data(c) for easy visualisation. Yellow boxes denote burn scars which are correctly labelled or unlabelled. White boxes denote mislabelled/misidentified burn scars. The defects are attributed to no segmented labels available and negligible instances of these features in the data-set.}\medskip
  \vspace{-0.5cm}
\label{fig:defect_results}
\end{figure*}
For burn scar segmentation task, we base our network on the UNet\cite{ronneberger2015u} architecture, where the feature activations from encoder were stored and transferred to corresponding decoder layer for concatenation.

\textbf{Encoder} The encoder network consists of 3x3 convolution layers along with batch normalization layers, ReLU non-linear activation layers, and 2x2 max-pooling layers.

\textbf{Decoder} The decoder network consists of UpSampling layers, which performs 3x3 Conv2DTranspose, 3x3 Convolution along with batch-normalization layers and Dropout2D layers with a dropout value of 0.1. The output results are thresholded to obtain a binary output map denoting the burn scars.

\subsection{Datasets}
This dataset consists of a visible and near infrared satellite imagery from LANDSAT8 of the Amazon Rainforest. The dataset was acquired for 2017 and 2018 from over 4 states, namely \textit{Tocantins, Maranhao, Mato Grosso, Para} covering over four terrestrial Brazilian biomes namely \textit{Cerrado, Amazonia, Caatinga, Pantanal}. It consists of 299 samples VIS-NIR image pairs of size 1024x1024 with ground truth, which are binary images, in which 1 represent burn scars in the forest and 0 are areas that were not affected by the fire.

The dataset can be visualised in Fig \ref{fig:problem_description}. As can be seen the ground truth is noisy and also partially labelled, sometimes mislabelled  as can be seen in Fig \ref{fig:problem_description} (a). The dataset was curated by National Institute for Space Research (INPE) as part of the Quiemadas Project\cite{INPE_queimadas_2018}.

\subsection{Inference and Training}
The output map is subjected to a binary crossentropy loss which is backpropagated to train the AmazonNet model in an end-to-end fashion. Concatenated $[\textbf{x}^i_{RGB}, \textbf{x}^i_{NIR}]$ is fed to the network. Adam Optimizer with a starting learning rate of 0.0001 is used for minimizing the loss function. The model was fine-tuned for 50 epochs. The batch size of the training datasets was eight whereas, for validation datasets, the batch-size of 4 was chosen. Inbuilt callbacks functions, namely EarlyStopping, ReduceLROnPlateau \& ModelCheckpoint were used for training our model.%
%

\section{Results}
The model obtained a training accuracy of 69.51\% \& a validation accuracy of 63.33\%. The results are presented in Fig \ref{fig:problem_description} (f) \& Fig \ref{fig:results} validating the efficacy of the utilising U-net based segmentation in burn scar identification. As can be seen in Fig \ref{fig:results}, the network correctly identifies unlabelled fragmented burn scars (denoted as yellow-dash boxes) and deselects wrongly labelled areas (denoted as white-dash boxes) in the output binary map (correctly labelled outputs are highlighted as yellow and deselected labels as white).

Our network, however, fails to distinguish river and cloud patterns from burn scars as can be seen in Fig \ref{fig:defect_results}. Defects emerge when our network segments (a) river (b) meanders and ox-bow lake and (c) clouds as burn scar patterns. 

It is interesting how in sample 2 and 3 in Fig \ref{fig:defect_results}, the network accurately segments the small fragmented burn scars but absolutely fails to reject these. This can be attributed primarily to (i) lack of any labelled examples and (ii) negligible samples containing the above geographical features in the dataset.

\section{Conclusion and Future Work}
We utilised a partially/mis-labelled dataset representing burn patterns in Amazon rainforest to propose U-net based segmentation network to correctly identify burn scars \& reject incorrect labels, demonstrating the effectiveness of AI in fragmented burn scar identification. We presented shortcomings \& consider resolving these as future work.

\section*{Acknowledgements}
Authors thank Prof. Subhasis Chaudhuri, IIT Bombay \& Paulo Fernando Ferreira Silva Filho, Institute for Advanced Studies, Brazil for discussion and productive comments. This work was partially completed as part of TechForSociety initiative at Koloro Labs. Satyam Mohla and Sidharth Mohla acknowledge support from Microsoft for AI for Earth Grant \& Shastri Indo-Canadian Institute for SRSF research fellowship.

\bibliographystyle{IEEEtran}
\large{\bibliography{root}}

\end{document}